# Iranian License Plate Recognition Using a Reliable Deep Learning Approach

Soheila Hatami[1], Majid Sadedel[2*] and Farideh Jamali[3]

[1]Department of Mechanical Engineering, Tarbiat Modares University, Tehran, Iran. (E-mail address: soheila.hatami@modares.ac.ir)

[2*]Department of Mechanical Engineering, Tarbiat Modares University, Tehran, Iran. (Tel: +98(21)82884987, Fax: +98(21)82884987, E-mail address: majid.sadedel@modares.ac.ir) – Corresponding author

[3]Department of Mechanical Engineering, Tarbiat Modares University, Tehran, Iran. (E-mail address: f.jamali@modares.ac.ir)

## Abstract

The issue of Automatic License Plate Recognition (ALPR) has been one of the most challenging issues in recent years. Weather conditions, camera angle of view, lighting conditions, different characters written on license plates, and many other factors are among the challenges for the issue of ALPR. Given the advances that have been made in recent years in the field of deep neural networks, some types of neural networks and models based on them can be used to perform the task of Iranian license plate recognition. In the proposed method presented in this paper, the license plate recognition is done in two steps. The first step is to detect the rectangles of the license plates from the input image. In the second step, these license plates are cropped from the image and their characters are recognized. For the first step, 3065 images including license plates and for the second step, 3364 images including characters of license plates have been prepared and considered as the desired datasets. In the first step, license plates are detected using the YOLOv4-tiny model, which is based on Convolutional Neural Network (CNN). In the next step, the characters of these license plates are recognized using Convolutional Recurrent Neural Network (CRNN), and Connectionist Temporal Classification (CTC). In the second step, there is no need to segment and label the characters separately, only one string of numbers and letters is enough for the labels. The average response time for the License Plate Detection (LPD) part of the proposed method is 0.0074 seconds per image and 141 frames per second (fps) in the Darknet framework and 0.128 seconds per image in the TensorFlow framework. Also, this method with a limited number of data was able to provide a model with a minimum error and low volume space for storage with less than 2MB for the Character Recognition (CR) model and an average accuracy of 75.14% for the end-to-end process. The average response time for this end-to-end proposed method is 0.435 seconds per image.

## 1. Introduction

Automatic License Plate Recognition (ALPR) has abundant applications all over the world such as controlling urban traffic, recording violations on the streets and roads, finding stolen vehicles, paying tolls, parking entrances, and many other ones, which have led to the development of smart cities. Due to its wide range of applications, this system has confronted many challenges, some of which have not been resolved yet.

A usual ALPR system consists of three main steps: License Plate Detection (LPD), Character Segmentation (CS), and Character Recognition (CR). Today, in addition to image processing algorithms, deep learning methods and neural networks are used to design ALPR systems. These concepts have led to significant progress in the field of ALPR. Unlike the traditional methods, the new ones do not segment license plate characters, so CS does not affect CR, and the algorithm becomes more adaptive [1].

According to many challenges in this field, some of them are related to the dataset. When taking an image of a license plate, some problems may occur in that the trained network could not locate the license plate in the image correctly. Problems such as the position or number of license plates, image quality, the angle of the license plate in the image, the existence of an obstacle to detecting the license plate, or detecting objects similar to the license plate are some of them. Also, this image can be taken in adverse weather conditions, which causes blurring and low quality. Moreover, the fact that an image is captured at which time of the day is very effective in determining its quality and could change the accuracy. Therefore, to overcome the mentioned challenges, it is necessary to prepare a comprehensive and diverse dataset of license plates.

Here, the main purpose of this paper is to present an approach which has a better accuracy/real-time performance trade-off to overcome some of the above-mentioned challenges. This method considers all steps of the ALPR process in terms of the algorithm and the dataset by using Deep Neural Networks (DNNs). The main contributions of the proposed method in this paper are as follows:

1. Due to the lack of a dataset compromising Persian letters and numbers written on Iranian license plates as string labels, not with separate labels for each character, and also the lack of a suitable dataset available for LPD in the input image, one of the main purposes of this paper is to first, provide suitable datasets for each part of the ALPR process that could be publicly available for other future works.
2. Also, many of the previous methods are used only for certain situations especially because of their datasets. For example, they do not have some types of license plates (such as political license plates). Therefore, there is a need to further generalize these methods and having different types of license plates in different conditions is one of the features of the presented datasets.
3. This proposed method for the CR step does not need the CS step and uses only a Convolutional Recurrent Neural Network (CRNN), and Connectionist Temporal Classification (CTC) to construct a structure and thus a cost-effective model for the ALPR process which needs low storage space and is suitable for mobile applications.
4. Also, in the final model, the license plates in the image are recognized in just 0.435 seconds after the image is taken by the camera. Of course, to detect a license plate, this value reaches 0.0074 seconds in the Darknet framework. The characters are then recognized in the TensorFlow framework in an average time of 0.127 seconds.
5. The frames per second obtained for the LPD in the Darknet framework is 141, which is much higher than previous models (at the of manuscript submission) and is suitable for real-time applications.

The remainder of this paper is organized as follows. In section 2 we briefly review the backgrounds of the existing datasets and also deep learning approaches and finally some of the related works. Details of the proposed dataset and the method are given in section 3. The evaluations and final results are presented in section 4. Finally, section 5 summarizes our conclusions.

## 2. Background and Related Works

### 2.1. Background

*Datasets*

In most of the issues discussed in the field of computer vision, the lack of a suitable dataset that covers all aspects of the work and has a good variety is one of the serious challenges in this area. The word "suitable" refers to the fact that the issue of bias in datasets has recently become a popular and controversial topic in such a way that it has been investigated regarding the major ALPR datasets [2]. With the view that the problem of LPD as the first step of the ALPR process is similar to the object detection from different points of view, and also given that the successful object detection networks are trained based on rich datasets such as ImageNet [3], COCO [4], and PASCAL VOC [5], so a proper dataset is required to train a LPD network. For this purpose, several datasets of license plates from different countries have been made available to the public, among them, datasets of Chinese license plates called CCPD [6] and Brazilian license plates called SSIG-SegPlate [7] have been used more.

Given that each country has its license plate, which differs from the license plates of other countries in terms of size, shape, and appearance, it is necessary to use the dataset belonging to the same country to train the networks. Therefore, due to this fact, to recognize Iranian license plates, a dataset consisting of Iranian license plates is needed. Compared to other countries' datasets, there are few datasets available for Iranian license plates. The dataset used in [8] has about 15,000 images that have been captured from three different regions. According to the information given in the paper, it has a good variety and all different environmental conditions are considered. This dataset does not have one of the most important requirements of a real-world dataset, being public, but according to the samples published in the article, the focus of this dataset is on one license plate per image, so this issue causes the trained network becomes weak in detecting multiple license plates per image. The dataset prepared in [9] is not publicly similar to the dataset of the previous one. Since the camera is fixed with a specific angle and position to collect license plate images, these images are not from different viewpoints and angles. The purpose of this fixed angle was also to capture one license plate per image, which again reduces the consistency of the dataset. Iranis [10] is the only dataset that has been made publicly available, but this dataset is appropriate for the last step of ALPR, i.e. CR. Unfortunately, this dataset treats characters as separate objects and labels them individually which is a tedious task to create a dataset. IR-LPR [11] is a multi-purpose dataset that has samples for both LPD and CR tasks. Among 20967 images of cars, 27745 license plates can be scanned for CR. For CR issues, this public dataset has annotations for each character, not the whole license plate.

*Deep Learning Approaches*

In recent years, Convolutional Neural Networks (CNNs), with their well-constructed algorithm, have made significant progress in computer vision and therefore they have become one of the most widely used networks in this field. Over time, various architectures of CNN have been proposed, each of which performed better than the others for a particular problem. The changes that each of them has compared to the others mainly include structural reformulation, regularization, parameter optimizations, etc. The depth of the network is especially a key factor in the development of its various architectures [12]. Regions-based Convolutional Neural Network (R-CNN) [13], You Only Look Once (YOLO) [14]–[17], and Single Shot MultiBox Detector (SSD) [18] are the best agents of CNNs.

YOLO [14] is the most offered object detector which is based on state-of-the-art algorithms. It is remarkable in accuracy and efficiency and is suitable for real-time tasks. The process of YOLO is that it divides the entire image into grid cells with equal sections and predicts several regulated bounding boxes and class probabilities for each section. Each of the bounding boxes that has a higher-class probability than the others are selected and the rest are ignored. YOLO consists of several versions, each of which has improvements in accuracy and performance compared to previous versions. YOLOv2 [15] can detect over 9000 object categories that are implemented through the Darknet-19 framework. Its architecture consists of 11 layers as the trained network on ImageNet (classification) and 19 layers on COCO (detection). The next version, YOLOv3 [16], uses the Darknet-53 framework. Its architecture consists of two sets of 53 layers. Compared to its previous version, it uses a residual block instead of anchor boxes and detects small images more accurately in terms of performance. Recently, another version of the YOLO detector, YOLOv4 [17], has been released, which is better than its predecessors in terms of accuracy and speed. The mean Average Precision (mAP) of the YOLOv4 improves by 10%, and also its frames per second (fps) by 12% compared to YOLOv3. SSD [18] only needs one shot to detect numerous objects. It can generate bounding boxes and detect the category of an object in one shot, which is faster than R-CNN in terms of processing.

Owing to the last step of the ALPR system, investigating one of the most remarkable studies which is the base for the CR step of this paper could be informative. CRNN with CTC [21] has been proposed for recognizing image-based sequences. This structure converts to a valuable approach that is a combination of Deep Convolutional Neural Network (DCNN) and Recurrent Neural Network (RNN) with CTC [22] in the transcription layer as the last layer and is superior to other methods in the task of image-based sequence recognition. For CRNN structure, at the underneath of the network, feature sequences are extracted from the input image by convolutional layers. The RNN is then placed on the head of the convolutional layers to predict each frame of the feature sequence, that is the output of the previous network. At the top of the CRNN and its output, a transcription layer is embedded to translate the predictions of RNN which are given as a matrix into a label sequence. Evidently, no character level annotations are needed for the learning procedure and sequence labels are enough as ground truth texts for the input images. Moreover, there is no limit to the dimension of the input sequences and the network can produce outputs in different lengths without preprocessing except for the height configuration of the input images.

## 2.2. Related Works

As mentioned, the ALPR process consists of three main steps and the focus of this paper is on the end-to-end of this procedure. Published studies on this topic have suggested different methods to implement the ALPR process. In this section, the variety of previous studies in the field of ALPR is reviewed.

Previous studies on ALPR are broadly divided into two approaches. The first approach involves traditional image processing methods such as texture-based, edge-based, color-based, and character-based approaches for the LPD step, pixel connectivity, projection profiles, and prior knowledge for the CS step and template matching techniques for the CR. The second approach that is more used today, involves modern techniques such as machine learning and deep learning algorithms.

Statistically, among the traditional methods for LPD, the edge-based method, or more specifically the edge density, has been widely used. For example, Massoud *et al.* [23] apply the Sobel edge detector filter to the image to detect the edges of the license plate. They also suggest a method for the CS step that uses the prior knowledge and information that exists in the size and font of the Egyptian license plate characters. Finally, the template matching method is used to recognize characters. The end-to-end accuracy of this system is 91%. Ingawale and Desai [24] propose a method to implement the LPD step that Canny edge detection is the basis of this process. The authors fulfill the CS step of this system but no detailed information is available. Finally, for the last step, namely, the CR step a primary neural network called Probabilistic Neural Network (PNN) is used, whose authors have an optimized preceding with other Optical Character Recognition (OCR) algorithms.

As expressed earlier, most recent works have used different types of neural network architectures to detect license plates. Due to the high features of CNNs in various applications, researchers have paid much attention to them. For an instant, Singh and Bhushan [25] employ Faster R-CNN based on two criteria of high accuracy and short run time, which uses the Region Proposal Network (RPN) for the concept of object recognition. The accuracy of the model to detect Indian license plates is reported to be about 99%. After applying a series of pre-processing methods on the cropped license plate image, Pytesseract (Python wrapper for Google's Tesseract OCR) has been used for both CS and CR steps. The end-to-end accuracy of the ALPR model is reported at 95%. Jamtsho *et al.* in their study [26] use the YOLOv2 to extract Bhutanese license plates. The proposed method achieves the overall mAP of 98.6% with 0.0231 training loss. The focus of this paper is only on the first step of the ALPR process. Instead of implementing each step of the ALPR procedure individually, Lin and Wu [27] set up a one-step model to recognize the license plate characters. They develop a method that uses the lighter version of YOLOv2, the YOLOv2-tiny, to extract license plates. The structure of the model changes, including removing the convolutional layer and changing the number of anchors and filters in some layers, which has an overall recall rate of 84.5% and a lower computational load than YOLOv2-tiny. Tourani *et al.* [8] relies on the benefits of YOLO, such as its real-time feature to recognize Iranian license plates, and their desired feature is more noticeable in the third version of YOLO, so they use YOLOv3 and they just change the last convolutional layer of it to fit this superior network on the Iranian license plate. The second YOLOv3 is used for both CS and CR steps, which means these two steps are combined to set up a classifier for recognizing all characters on the license plate. The end-to-end recognition accuracy of the proposed approach was 95.05% and the end-to-end average time to specify a sequence of characters in real-time is 119.73 milliseconds. Silva and Jung [28] base their study on a hierarchical CNN and use the same CNN for both steps, vehicle detection, and LPD. This network is inspired by YOLO architecture and built by making changes to its layers. This network is a FAST-YOLO type that is processed faster than typical YOLO and its computational cost is lighter. Also similar to the first two steps described earlier, another common CNN is used for both CS and CR steps, using the same FAST-YOLO architecture as the grid base, but with some changes to its architecture. Reducing the number of max-pooling layers from five to three, using transfer learning for the first eleven layers, cutting down the next layers after that, and adding four more layers to train them from scratch are some of the things that have been done to change the network architecture.

To deal with environmental challenges when localizing license plates in the image, Samadzadeh *et al.* [29] prefer a robust ALPR that also performs well in terms of execution speed. They use SSD and its MobileNetV1 for the first step of their study. For the CS step, they handle their method in two parts. For the first part, they use an open-source SOTA text detector as text detection and the second part involves two activation maps that they employ to correct the perspective transformation of the license plate and segmentation boundaries. Finally, the residual network, ResNet18, is used for the CR step. Evaluations show that the overall accuracy of the system (including the accuracy of the three main steps) is 95%. Also, the overall performance of the system is about 15 fps. Wang [24] also implements the license plate extraction step by SSD. Then for the CS, he chooses vertical projection methods due to the structure of license plates and at the end point, a CNN is preferred for the CR step. Wang *et al.* [30] propose a method that is more commonly used for face detection. They adapt the Cascaded CNN method named Multi-Task Convolutional Neural Network (MTCNN) for LPD by creating a series of reinforcements. The layers of this network are divided into three specific sets under the headings of Proposal Network (P-Net), Refine Network (R-Net), and Output Network (O-Net) and each section performs a task. The P-Net results are trained by the R-Net subnet and then have the same task as P-Net with a distinct structure. The O-Net also has the same functionality as the previous two subnets and differs only in structure. Using the object recognition algorithm, the obtained regions are scored and the candidate with the highest score is selected as the license plate in the image. It is noteworthy that the method propose in the paper does not have a CS step, and after LPD, the license plate characters are recognized directly, for which a CRNN with CTC is used. The performance indicators of this system with data augmentation are such that its AP is equal to 98.8% and its performance is 64 fps. In [31] Silva and Jang, previously mentioned in another study, introduce the Warped Planar Object Detection Network (WPOD-NET), for LPD using YOLO, SSD, and Spatial Transformer Networks (STN), which has an average accuracy on different public dataset of 89.33%. The main difference between this network and the cases mentioned is in an affine transformation. This feature converts defective license plates that have lost the original shape of a license plate, to a rectangular shape or the same license plate format. In contrast to most research in this area, Al-Batat *et al.* [32] placed vehicle detection as the first step of a YOLO-based pipeline for ALPR. For vehicle detection, they have employed YOLOv2, and for LPD and CR, they have employed YOLOv4-tiny. Their approach was evaluated using five famous public datasets with an average accuracy of about 90.3%. In addition, this study added the capability of classifying vehicle types (emergency vehicles and trucks), utilizing ResNet50.

## 3. The Proposed Method

In this section, the proposed method will be discussed in detail. Due to the need for a dataset as the first step to detect license plates and recognize their characters, initially, both datasets for each of the steps (LPD and CR) are prepared. After preparing the datasets, it is necessary to detect the rectangles of license plates in the input image or video frames, and for this reason, a new model called YOLOv4-tiny which is a lighter model of YOLOv4 with fewer parameters that helps in its real-time application by sacrificing accuracy is used. Since the CR step is performed in the TensorFlow framework, by using CRNN and CTC structure, and the LPD step is performed in the Darknet framework, it is not possible to connect the Darknet and TensorFlow frameworks to each other. Therefore, to connect these steps, the weights obtained from the training step of LPD in the Darknet framework must be converted into a model within the TensorFlow framework with the method described in [33]. After converting the weights, the CR step trained in the TensorFlow framework is attached to this model, and as a result, with this connection, the whole process is performed in this framework. As Figure 1 shows the whole process in a flowchart, the license plates are detected, then these license plates are cropped from the input image, and finally, the characters written on the license plates are recognized.

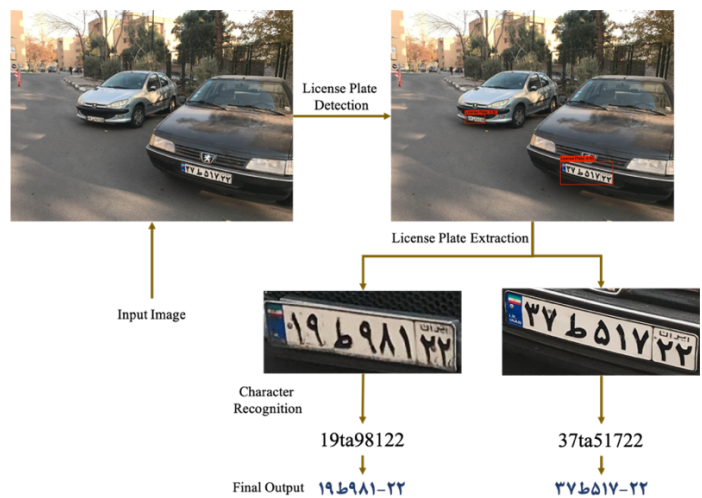

**Figure 1.** The overall diagram of the proposed method

### 3.1. Dataset for License Plate Detection

In general, the dataset used for this section contains 3065 images, which due to security issues, these images are not available to the public, except for people who need these images for their research works and should not use any commercial use of these images. According to these conditions, 1466 images from this dataset were purchased from a website and another 1600 images were prepared under different conditions. Later, as mentioned, these 1600 images could be made available to other researchers.

1466 images of the dataset which were purchased taken from a close distance to the car and also only one license plate is labeled in each image. The rest of the images collected in the dataset, 1600 images, have a much greater variety. Weather and lighting conditions, the distance of the camera from the car, the view angle of the license plates, and the use of cameras with different qualities are all factors that have been taken into account in preparing the dataset. Also, some of these images were taken from different sources and places; using multiple car sales websites (people with different views have taken pictures and this helps to diversify the dataset), in parking lots, parked and moving cars, at different times of the day and with different lighting of the day (like sunny and cloudy weather) and also with a distance of 1 to more than 10 meters from cars as Figure 2 shows. The reason for this amount of diversity is the learning and familiarity of the proposed network with different scenarios of images.

Dataset images are manually labeled by the labelImg toolbox. Most of the images were taken using mobile phone cameras with different resolutions and formats, but in the end, for convenience, all these images were saved in .jpg format. Finally, all the labels of these images were converted to .txt file format, which could later be used by models such as YOLO.

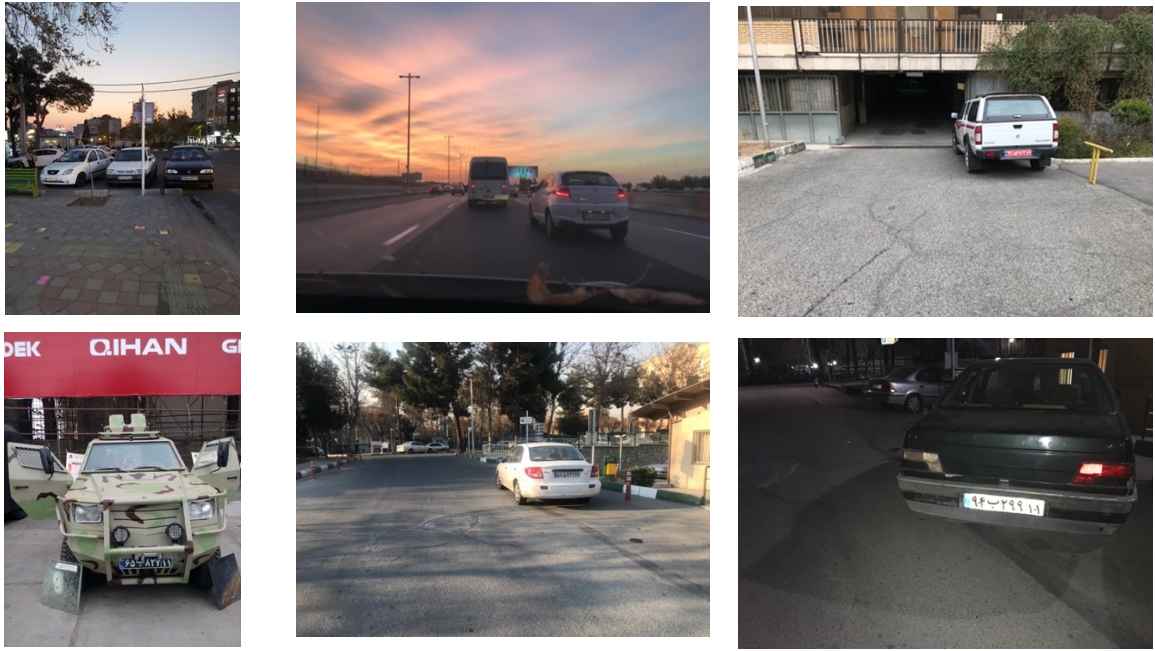

**Figure 2.** Examples of LPD dataset images

### 3.2. Dataset for Character Recognition

To recognize Persian characters written on license plates, a dataset including Persian license plates in which all characters are labeled has been purchased and used. In general, this dataset has 5455 images, of which 455 images have blank labels and 818 images have a number of characters less than 8 values. The reason for assigning some images that do not have the characters of the license plate and therefore have blank labels, is to prevent the recognition of different objects as characters. For example, it may recognize a picture of a fence or any striped object as the number 1111, and this is necessary to strengthen the model. But there is no need to bring these images here, so they have been removed from the dataset.

In this dataset, all image labels are in .json file format. In these label files, the position of each character is given separately, which does not require the position of these characters in this proposed method. For this reason, we have converted all these Persian characters into English equivalents, and finally, we have a string of English numbers and letters for each image. Now all these images, the number of which reaches 3364, are given to the desired network for training. In this case, almost all types of data exist; single-digit, single-letter, as well as a combination of numbers and letters, are all present in this dataset. These labels are considered image names after being prepared in a string format, so there is no need for separate label files next to the images. The format of all images is the same as .jpg.

In this dataset, all the characters of Iranian license plates are covered as follows:

{alef (الف), be (ب), pe (پ), te (ت), se (ث), jim (ج), dal (د), ze (ز), sin (س), shin (ش), sad (ص), ta (ط), eyn (ع), fa (ف), qhaf (ق), kaf (ک), gaf (گ), lam (ل), mim (م), non (ن), he (ه), vav (و), ye (ی), tashrifat (تشریفات), malol (معلول), D, S}

It also includes all numbers from 0 to 9.

Given that Persian letters cannot be considered exactly as a label for an image, so it is necessary to first consider them equivalent to English letters and numbers, and then after their final stage, return them to the desired Persian letters. As shown in Figure 3, the number of characters used in the dataset is shown, and for numbers except 0, the number of each number is approximately the same. Of course, in some cases, such as the numbers 1, 8, and 9, due to a large number of these numbers in real-world license plates, here, too, are more than other numbers. Also, the letters such as "sin" or "be" are less here due to their small number on typical license plates. At the same time, some license plates, such as "tashrifat" license plates or special license plates, are much less commonly seen on the streets than other license plates, so there are fewer numbers of them in this dataset.

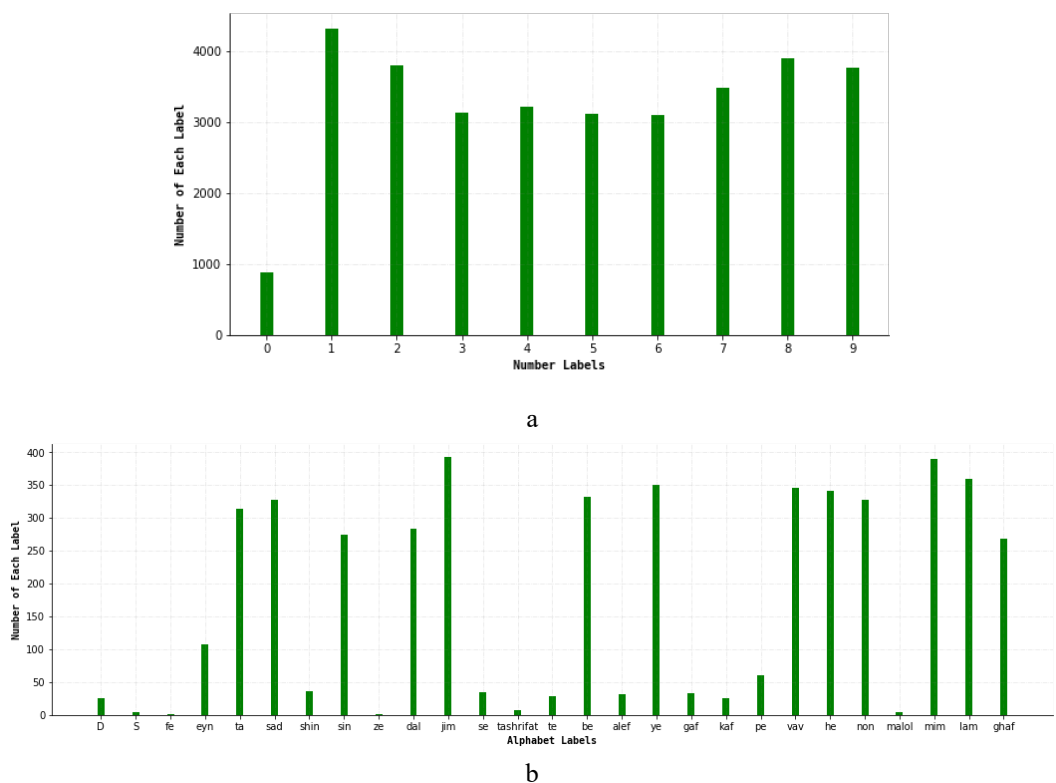

**Figure 3.** The number of different labels used in the CR dataset: a) Numbers b) Words and letters

Here, according to the available dataset and the selected network model, 80% of the images, i.e., 2692 images for the training, 10%, i.e., 336 images for the validation, and the last 10% part for the test set are considered. After preparing the images for the dataset, the license plates need to be detected and then recognized in two steps, respectively.

### 3.3. License Plate Detection Network

In this section, the YOLOv4-tiny model is used to detect license plates in the input image and also evaluate the first dataset. Considering that this model has been implemented and trained in the Darknet framework and belongs to it and gives the best outputs, this framework is used first for training and even testing, but since working with this framework is less flexible, after the LPD training process in this framework, the weights are converted to a model that can be used in the TensorFlow framework. So, we can easily connect the converted model to later parts that are not in the Darknet framework, for example in the TensorFlow or even PyTorch frameworks.

In this model, all input images are converted to 416 × 416 dimensions and considered as input for the training network. Also, in this architecture the same default configuration is used for YOLOv4-tiny, only in some cases changes have been made. Among other things, due to the limited memory of the GPU used, the batch and subdivisions are changed to 32 and 4, respectively. The number of 32 batches means that the data is selected as 32 images (one batch) for processing, and a high value requires more memory for the GPU. Also, the images of each subdivision are converted into 4 blocks and used in parallel for processing. The momentum remains the same as 0.9 and is suitable for stabilizing the gradient. The weights decay value is the same as the default value of 0.005, which updates the weights weaker for normal features and also eliminates imbalances in the dataset. The learning rate value remains unchanged at 0.00261, which is a relatively proper value for this structure, and its low value prevents the network from diverging due to unstable gradients. The number of iterations in this model is equal to the value of 10000. Given that the images for training are 2452 (80% of the dataset images) and the data are trained in groups of 32, we need about 76 iterations for each period of seeing the total data, which means that it is intended for 10000 iterations, about 132 epochs are calculated (132 times the data is viewed from the beginning to the end) and this is a relatively suitable amount because a small dataset requires at least 100 epochs for training. In this case the remaining 20% of the images in the dataset are divided into two groups for the validation and test set (each comprising 306 images). Also, in this structure, the same default data augmentation is used, and some of its techniques such as Mosaic are not suitable for this model of data and it reduces the average accuracy. Finally, due to having only one class, the number of classes has changed to 1, and due to the change in the number of classes, the filters also have changed, and (number of classes + 5) × 3 gives a result of 18 for the filters. Since the weights trained in the Darknet framework need to become a model in the TensorFlow framework, their anchor boxes should not change and should remain the same. Finally, the threshold value for IoU is set to be 0.50.

To train more and better the data, pre-trained weights on the COCO dataset are used. These pre-trained weights lead to better and faster network learning and the transfer of learning through them. According to the previous explanation, after training the data in the Darknet framework, the weights obtained are transformed into a model in the TensorFlow framework. So, we can connect this step with the next one which is also in the TensorFlow framework.

In the end, the obtained weights in the Darknet framework have 23.5MB size and the model in the TensorFlow framework has 26.3MB size in storage space, which there is not any significant difference and both of them need a little space to be stored.

### 3.4. Character Recognition Network

To recognize the license plate characters in the proposed method, it is no longer necessary to segment the characters individually, and as mentioned before, it is only necessary to detect the license plate and obtain its rectangle, as the input for the recognition step.

Before addressing the desired structure for training, it should be noted that given that typical Iranian license plates have eight characters, in images that have the number of characters less than eight, this is compensated when the model is implemented with the "X" character. Wherever, this character is, some parts of the license plate characters do not exist (this is true for those license plates which have eight characters, because for example "tashrifat" license plates have less than eight characters). As an example, this character may be seen twice at the time of prediction, i.e., the model has recognized only six characters and has not seen other characters, because if it fails to recognize otherwise, it says no detection. The trained model may in some cases receive, for example, eight characters (or less) but can only recognize some of them, in which case the term "Not-Known" refers to its non-recognition.

In this proposed method, by adopting the overall structure described in [21], a CRNN, and finally a CTC within the transcription layer, are used for CR. In this part, the CNN is used to extract the features of the input image (with the width and height, 200 and 50, respectively), this network extracts a sequence of feature vectors. The RNN then propagates the information during this sequence, and its output is a matrix with numbers (probabilities) that assign a value to each element of that sequence. Because of having a sequence of arbitrary lengths and also, overcoming the vanishing gradient problem, Long-Short Term Memory (LSTM) as a type of RNN is selected. Hereby, for this case two LSTMs are combined as bidirectional LSTMs to interpret contexts from both directions (one forward and the other backward). Now, by using this matrix, which encodes the input, the amount of loss must be calculated to train the neural network with Adam optimizer, and its information can also be decoded to obtain the text written in the input image. Both expressed tasks are performed by the CTC in the transcription layer. Finally, the CTC recognizes the characters. It is sufficient to give the loss function of the CTC, the ground truth label related to the input image, and the output of the RNN matrix to complete the task. So, in this case, the position and width of the characters no longer matter. There is also no need to over-process the recognized characters.

As shown in Figure 4, the CNN sees the image horizontally and thus extracts its features. Each of these horizontal positions is called a time step. After the output of the matrix is obtained, the values of this matrix, together with the ground truth label of each input image, are given as input to the loss function of the CTC in the transcription layer. This operation does not need to know where each character occurs, but considers all the different states of the ground truth label related to the text. Then the probability of each time step occurring with the probability of other time steps occurrence multiplies (based on the ground truth label). It then adds together all the states that are intended for one input, and the result of any set that is greater than the others is considered as output.

Also, to decode duplicate characters, the "blank" character is used, which is marked with a "-" and will be deleted at the time of decoding. Its function is that if two characters need to be repeated, this symbol is used among them, and at the time of decoding, this action leads to the recognition of duplicate characters. All of the above are for when the network is training and needs to minimize loss. The sum obtained from different states of input text is the amount of loss that can be expressed as follows:

$$Loss\ Function = -\sum_{I_i, l_i \in X} \log p(l_i|y_i) \text{ which } X = \{I_i, l_i\}_i \qquad [21]\ (1)$$

In this regard, X is the dataset, $I_i$ is the training image, $l_i$ is the ground truth label, and $y_i$ is the sequence created from the input text image by the CNN and RNN. If the network predicts the data that it has not seen before, there is no longer a ground truth label. In this case, it should use the best path decoding. In this type of decoding, the best path is selected in such a way that in the output matrix of the RNN, at each step when the probability of occurrence of the predicted character is high, the same character is placed in the path and this process continues. Find that until the prediction is done, then the duplicate characters are predicted and finally the "blank" character is removed. For this reason, the greedy search decoder is used in this section, but for more complex texts, the beam search decoder can be used. In the beam search decoder, candidates are first introduced using a tree structure, then points are given to each using the best path, which not only increases the accuracy but also the time for prediction much longer, and this is a drawback for real-time and high-speed methods.

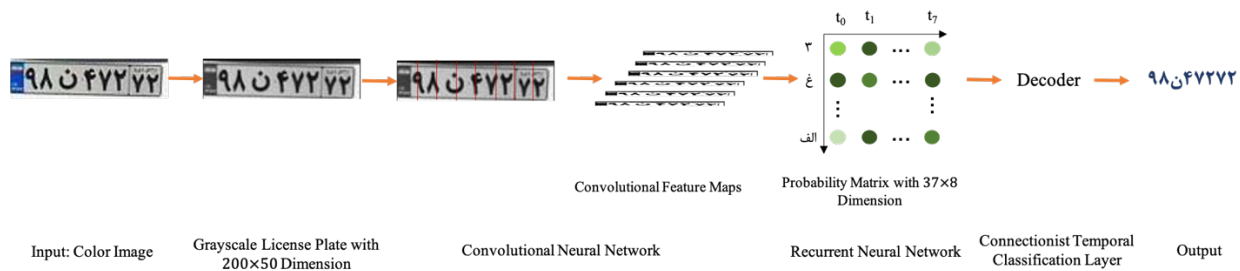

**Figure 4.** Overall diagram of the CR

Table 1 clearly shows the general structure of the network used for the CR step. Using this structure for CR provides a model with less than 2MB, which is very small in size. Also, in the end, the number of parameters of the model used is 434,918, which is a relatively small value and can be used for mobile devices.

Table 1. Summary of network structure used for CR

| Type | Configurations |
|---|---|
| Input Image | 200 × 50 Gray scale |
| Convolution layer | Units: 32, Kernel size: 3 × 3, Activation function: relu, Kernel initializer: he_normal, Padding: same |
| Maxpooling | Pool size: 2 × 2, Stride: 2 |
| Convolution layer | Units: 64, Kernel size: 3 × 3, Activation function: relu, Kernel initializer: he_normal, Padding: same |
| Maxpooling | Pool size: 2 × 2, Stride: 2 |
| Dense layer | Units: 32, Activation function: relu |
| Dropout | Rate: 0.2 |
| Bidirectional LSTM | Units: 128 |
| Dropout | Rate: 0.25 |
| Bidirectional LSTM | Units: 64 |
| Dropout | Rate: 0.25 |
| Dense | Units: 38, Activation function: softmax |
| Transcription | - |

All calculations are performed in the Ubuntu 20.04 operating system environment. The hardware used is Intel Core i7 3.50GHz CPU, 32GB RAM, DDR4 for CPU, and Geforce GTX 1050 4GB for GPU, cuDNN, CUDA, TensorFlow 2.3, and Python 3.9 are also used.

## 4. Results

### 4.1. License Plate Detection

Since the main framework for implementing the YOLOv4-tiny model is the Darknet framework, therefore, by using this framework, more speed and accuracy can be obtained. But as mentioned before, this framework has less flexibility and as a result, after training the data in this framework, the trained weights need to be converted into a model in the TensorFlow framework and thus that model can be used for further processing.

The training data, i.e., 2452 images, are given to the YOLOv4-tiny network for training, and according to the network framework, the training time for this number of images is less than 90 minutes. Validation data is then used to check the network output on the images, and finally, the weights obtained are used to evaluate the model on images and video frames. According to Figure 5, which is stored by the network during training, the network loss is decreasing with passing iterations. So, this decrease has reached dramatically less than 1 even before 1000 iterations, and after that, this decrease has continued.

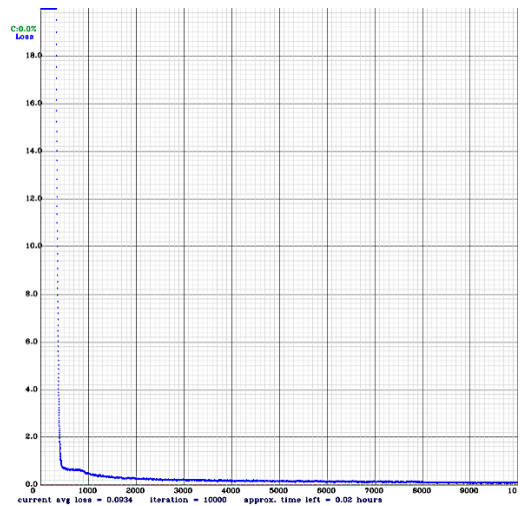

**Figure 5.** The network loss with to its iteration number

Given that the images in the validation set are different from the test images, the final results are just performed on the validation images. Figure 6 shows examples of these detections in different situations. As it turns out, these images are in different situations: night, day, with more than one license plate and different distances from the camera. The trained network gives as outputs a rectangle drawn around the license plate, along with the name of the class (License Plate), the network reliability of the detection performed, as well as the coordinates of the detected rectangle (including the top, bottom, left and right points relative to the whole image).

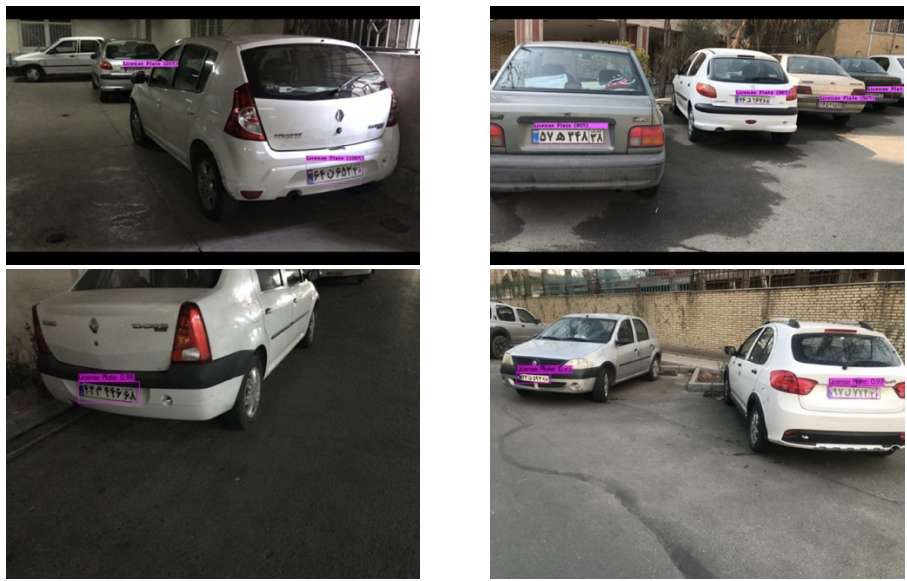

**Figure 6.** Examples of LPD using the Darknet framework

Finally, for our detection, we have the correctly detected license plates as True Positives (TP), not detected license plates as False Negatives (FN), and other items known as license plates while they were not, called False Positives (FP). For the case where the license plates may not be detected, the user needs to check all the data and there is no way to warn the user. But for the case where another element is mistakenly detected as the license plate, given that this detected element enters the next step, i.e., CR, in this step, the characters are not recognized correctly and the user only needs to check the relevant image. As can be seen from Table 2, the number of license plates that have been correctly detected is much higher than the number of license plates that have not been detected or even mistakenly considered other elements as license plates.

**Table 2.** Number and distribution of detected license plates in each set

| Detection type | Training images | Validation images |
|---|---|---|
| TP | 2255 | 539 |
| FP | 226 | 50 |
| FN | 315 | 105 |

As it is clear, Figure 7 includes the results of the images taken after converting the weights trained in the Darknet framework to a model in the TensorFlow framework. As it turns out, this model can easily detect more than one license plate in the image, and can also detect images containing license plates with different conditions. The converted model in this framework also gives as output the name of the desired class (License Plate), the network reliability of the detection as well as the coordinates of the detected rectangle (including its top, bottom, left, and right points relative to the whole image). The coordinates of the detected rectangle are used to crop the license plate from the whole image to recognize its characters.

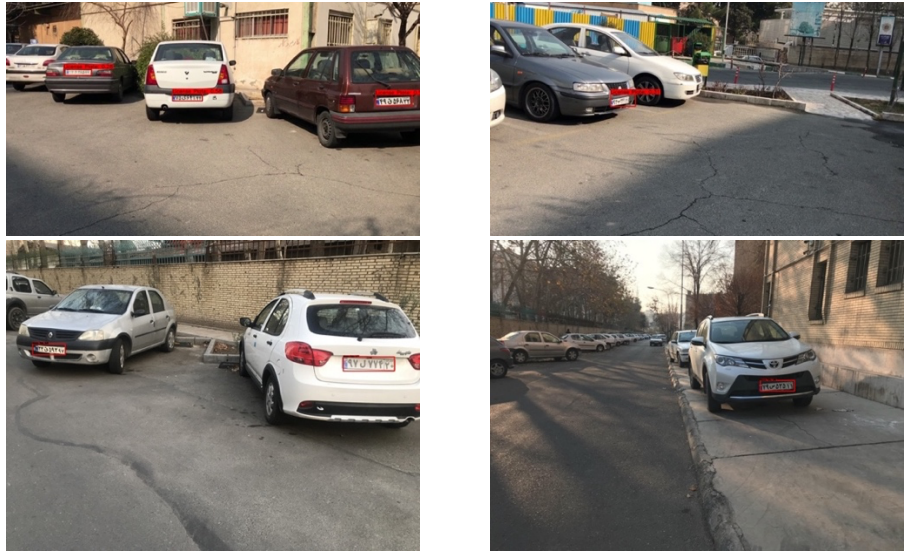

**Figure 7.** Examples of LPD using the TensorFlow framework

As can be seen from Table 3 and Table 4, this method has reasonable results for the Darknet framework, it can be used for real-time applications. However, due to the fact that the main environment for implementing YOLO models is not the TensorFlow framework and requires more optimization, so the results have the same accuracy but lower speed. At the same time, the model obtained in the TensorFlow framework can be easily connected to models in other frameworks and this is a big advantage.

### 4.2. Character Recognition

To obtain a suitable model in the TensorFlow framework, it is first necessary to consider the training images for the training process. These images are all in the form of images that have labels of numbers and letters. Due to the structure of the networks used for this model, its training time is about 30 minutes, and this time is such that only a string of numbers and letters is needed and there is no need to segment the characters. Figure 8 shows examples of CR for some Iranian license plates with different characters. At the top center of these images, a sequence of English characters, after which, only the English letter of this sequence must be converted to its Persian equivalent. Besides, the ground truth for each one is added. As a result, at the bottom of each license plate is written a string of numbers and letters equivalent to the above prediction.

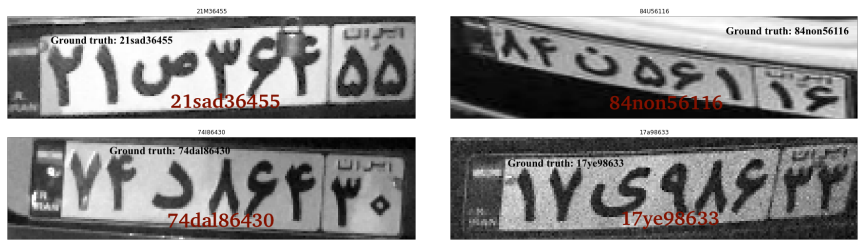

**Figure 8.** Examples of CR

### 4.3. End-to-End Process

Finally, the model obtained in the LPD step is merged with the model obtained for the CR step, and the results are obtained on the images and video frames in Figure 9.

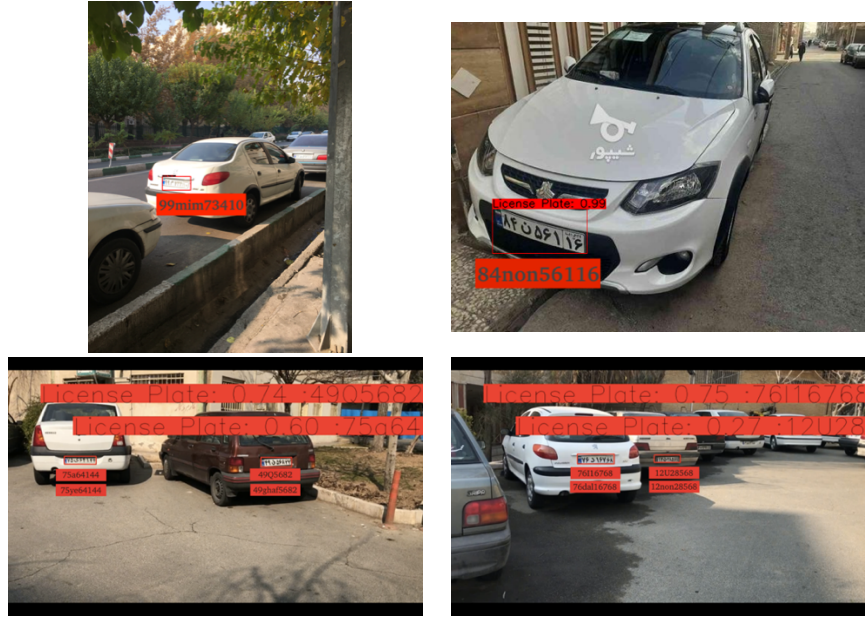

**Figure 9.** Examples of the end-to-end proposed method

In this case, as it turns out, the license plate rectangles are first detected using the model converted in the TensorFlow framework, and then the characters written on them are recognized without being segmented. Eventually, only the English letters of these characters converted to their Persian equivalents. For this reason, in addition to these predictions (for images and video frames), their predicted equivalents are also included. In fact, since the projection rectangle needed to be large for better visibility, part of the prediction was not included in the image. But to ensure the output, a text file has stored the output for each frame, and the output written under each license plate is the output of the text file.

### 4.4. Challenges of the Process

For each of the two separate steps of the ALPR process, there are some challenges, which are addressed below:

1) License plates may have a lot of dirt, including dust, or even a lot of tilting due to factors such as accidents. This will make the license plates illegible.

2) The license plate rectangle obtained as the output of the first step may not cover the entire license plate and as a result, the characters are less than usual.

3) Incoming images during filming may cover the blur or even just part of the license plate. Also, the camera view of the license plate may not be appropriate.

4) Some of the characters written on the license plate have many similarities to each other, and the low quality of the input image exacerbates this issue (illegibility and consequent misrecognition). For example, the numbers 3 and 4 may be mistaken for each other.

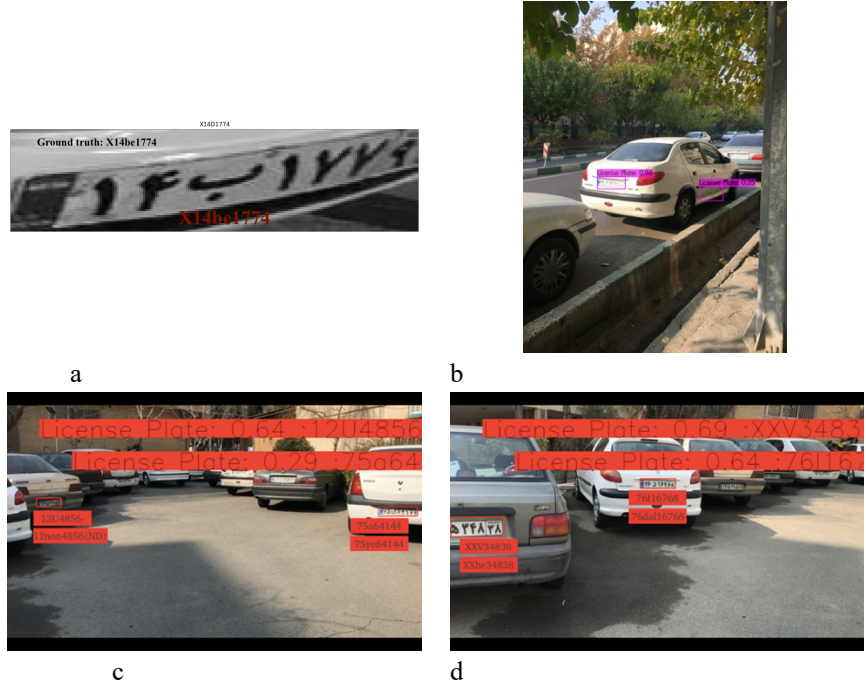

a   b

c   d

**Figure 10.** Examples of network errors

As Figure 10 shows, there are challenges for both steps in the proposed ALPR approach. In image a, the license plate character has been recognized less than eight, and character "X" indicates that the number of characters has a value of less than eight. Also, the number "4" is incorrectly recognized instead of the number "9". In image b, the license plate and part of the car are detected as outputs. In part c, for the Peugeot 405 car, the number "4" was incorrectly recognized instead of the number "2", and due to the incorrect license plate detection (location of the rectangular), the number "8" at the end of the license plate is not correctly located in the detection rectangle, so this number is not recognized correctly. Finally, in part d, which is part of the video evaluation, the number of characters for the Pride car is six. For example, in some cases, the license plates are dirty or the camera does not have a good view of them. In the end, for all the results to be comparable at a glance, they are all listed in Table 3. This table shows the average execution time for images with different qualities and also the frames per second value for videos. As the results show, this method is very suitable for images, and with this number of data has a significant accuracy. Table 4 shows these accuracies for different steps of this proposed method well.

**Table 3.** Execution time and frames per seconds results on image and video

| | Step | Framework | Execution time (sec) | | Step | Framework | Frames per seconds (fps) |
|---|---|---|---|---|---|---|---|
| **Image** | LPD | Darknet | 0.0074 | **Video** | LPD | Darknet | 141 |
| | | TensorFlow | 0.128 | | | TensorFlow | 7.8 |
| | CR | TensorFlow | 0.127 | | ALPR | TensorFlow | 2.3 |
| | ALPR | Darknet + TensorFlow | 0.255 (separate calculation) | | | | |
| | | TensorFlow | 0.435 | | | | |

**Table 4.** Average precision on training and validation images

| Type of data | LPD | CR | ALPR |
|---|---|---|---|
| | **Darknet** | **TensorFlow** | **TensorFlow** |
| **Training images** | 88.66% | 91.01% | 77.61% |
| **Validation images** | 87.81% | 87.22% | 75.14% |

## 4.5. Comparison of the Proposed Method with other Methods

As Table 5 shows, the proposed method has been able to achieve acceptable accuracy with this limited amount of data. This method has also been able to provide a compact model that consumes limited storage space. The methods presented in previous studies have datasets with large numbers of images, higher-volume models (storage space), or datasets of images with limited conditions: that is, there is only one vehicle in the images, for example, or images are taken from very close distance. These issues and some others have all contributed to the high accuracy of these methods. While in this proposed method, an attempt has been made to obtain the best results with the least conditions.

**Table 5.** Comparison of some recent methods with the proposed method

| Step | Method | Pros | Cons | AP & Time |
|---|---|---|---|---|
| LPD + CR [30] | CNN + CNN, RNN, CTC | Without character segmentation, use a dataset with more than 250,000 images and a model 15 times lighter than the YOLOv3 model | Use a highly computed chain network (with three sections) for LPD | 98.8% |
| LPD + CS + CR [8] | (Preprocessing) CNN (YOLOv3) + CNN (YOLOv3) | Using a dataset with more than 15,000 images (data with appropriate diversity), improving images using histogram equalization | Using preprocessing, using two YOLOv3 networks with high computational calculations, having only one car and one license plate per image of the dataset, a three steps process | 95.05%, 119.73msec |
| LPD + CS + CR [29] | CNN (SSD300) + (Preprocessing) Using CRAFT for detecting text blobs and characters + CNN | Using a method for CR with consuming less storage space | Using a dataset with a limited number of images, a three steps process | 95%, 66msec (per license plate) |
| CR [34] | (Preprocessing) CNN (YOLOv3) + Tesseract | Use small number of images in dataset to reach a good AP | Focus on one step (CR), use preprocessing, use a network with high computational calculations | 99.2% |
| **LPD + CR (The Proposed Method)** | CNN (YOLOv4-tiny) + CNN, RNN, CTC | **No need for CS, no preprocessing, a model with consuming less storage space, real-time application for LPD step, using small number of images in dataset to reach a good AP** | Time consuming of the CR step | **75.14%, 435msec** |

## 5. Conclusions

In this paper, a method for automatic Iranian license plate recognition was presented. In the proposed method, vehicle license plates are first detected in the input image or video frames using CNN, which is a type of deep neural network, then the characters written on license plates are recognized by CRNN and CTC structure. Initially, due to the lack of a suitable dataset for Iranian license plates, a dataset of 3065 images was prepared, which has a wide variety of images. For the next step, CR, a dataset of 3364 images including characters of license plates have been prepared. After that, the proposed method was performed by using YOLOv4-tiny model and in the Darknet framework for LPD. Due to the limitations of using this framework and the impossibility of connecting it to other frameworks such as TensorFlow for the next step, the trained weights were converted to a model in the TensorFlow framework. As the results show, the Darknet framework gives much better results. At the final step, this converted model connected to the model obtained in the TensorFlow framework which was created from the CR step, and a unified method without CS was available. Also, some main achievements obtained are as follows:

1. The small size of the proposed model (less storage space)
2. Short execution time (speed) of the proposed method
3. Obtaining a model with high average accuracy even with the limited number of images in the datasets

Considering that the use of YOLO models for the OCR step that has been proposed in previous works requires separate labeling of characters on license plates, for this reason, using other network models, for example, using a combination of deep RNNs can help to improve this part. Here we do not need to have the coordinate of each character and only a sequence of them is needed. As a result, this type of model is implemented in frameworks other than Darknet, and therefore converting weights to a model in the TensorFlow framework can be efficient.

In general, according to the results obtained in the results section and Figure 6,Figure **7**, andFigure **8**, as well as Table 5, it can be concluded that all the approaches of the article, have been achieved. In addition, the purpose of presenting a reliable model here is that by changing different inputs and in different conditions that the model is not familiar with, the proposed method can obtain desirable outputs. Because this demand has been fulfilled to an acceptable extent, a robust model has been obtained here.

In the future, the speed of the obtained model for testing data in the TensorFlow framework could be improved. It is also possible to expand the prepared dataset with more variety of images, and as a result, the performance of the proposed method can be improved. The method used for the final step of the process can also be used for other image-based sequence tasks.